# Domain-Robust Marine Plastic Detection Using Vision Models

Kataria, Saanvi

## Abstract

Marine plastic pollution is a pressing environmental threat, making reliable automation for underwater debris detection essential. However, vision systems trained on one dataset often degrade on new imagery due to domain shift. This study benchmarks models for cross-domain robustness, training convolutional neural networks - CNNs (MobileNetV2, ResNet-18, EfficientNet-B0) and vision transformers (DeiT-Tiny, ViT-B/16) on a labeled underwater dataset and then evaluates them on a balanced cross-domain test set built from plastic-positive images drawn from a different source and negatives from the training domain. Two zero-shot models were assessed, CLIP ViT-L/14 and Google's Gemini 2.0 Flash, that leverage pretraining to classify images without fine-tuning. Results show the lightweight MobileNetV2 delivers the strongest cross-domain performance (F1 ≈ 0.97), surpassing larger models. All fine-tuned models achieved high Precision (~99%), but differ in Recall, indicating varying sensitivity to plastic instances. Zero-shot CLIP is comparatively sensitive (Recall ~80%) yet prone to false positives (Precision ~56%), whereas Gemini exhibits the inverse profile (Precision ~99%, Recall ~81%). Error analysis highlights recurring confusions with coral textures, suspended particulates, and specular glare. Overall, compact CNNs with supervised training can generalize effectively for cross-domain underwater detection, while large pretrained vision-language models provide complementary strengths. Future work should explore hybrid strategies, such as small CNN backbones with foundation-model priors and domain-aware sampling, to combine high Precision with Recall across heterogeneous marine environments and reduce labeling burdens at scale.

## 1. Introduction

Plastic debris in marine environments has become pervasive and harmful, endangering wildlife and ecosystems. More than 8 million tons of plastic enter the oceans annually, resulting in the deaths of over 100,000 marine mammals and turtles, as well as over a million seabirds, each year, with an estimated $13 billion in global economic costs [8]. Locating and removing submerged plastic waste is a challenging and labor-intensive task, motivating the development of autonomous underwater vehicles and computer vision (CV) systems to aid in debris monitoring and cleanup [16]. Deep learning has shown promise in detecting marine debris, significantly improving identification accuracy in recent studies [3]. For instance, Hipolito et al. achieved a mean average precision of over 98% in detecting underwater plastic using a YOLOv3 model, although on a limited dataset [3].

However, a persistent challenge is domain shift: models trained on a particular underwater imagery dataset often face degraded performance when applied to images from different locations, cameras, or conditions [11]. Underwater images can vary widely in visibility, lighting color cast, and background (reef vs. open water), making it difficult for a model to generalize beyond its training distribution [11]. Prior works have noted that many marine debris image datasets are highly localized to specific environments, which limits the generality of models [12]. For example, Sánchez-Ferrer et al. found that a detector trained in a controlled tank environment performed poorly on real seafloor images, highlighting the need for more diverse training data to encompass various conditions [11]. Van Lieshout et al. (2020) and others similarly argue for diversifying water conditions, visibility, and locations in training imagery to improve robustness [12]. Recent surveys confirm that many state-of-the-art models achieve high accuracy on specific datasets (often >60–70% mAP), but their real-world efficacy depends on how well the training data represent the variability of marine settings [9].

Addressing cross-domain robustness is thus crucial for the practical deployment of marine litter detectors in the wild. Researchers have begun exploring domain adaptation and generalization techniques for underwater vision. Chen et al. introduced a continual unsupervised domain adaptation method for a custom object detector, aiming to improve detection across different domains [1] progressively. Saoud et al. (2023) proposed ADOD, an attention-augmented YOLOv3 that learns domain-invariant features through a residual attention module and an auxiliary domain classifier, resulting in improved detection in unseen underwater conditions [12]. These approaches show that explicit domain generalization strategies can enhance robustness.

On the other hand, the rise of large-scale pre-trained vision models offers an alternate path: models such as OpenAI's CLIP learn from 400 million image–text pairs. It can recognize visual concepts in a zero-shot manner via natural language prompts [10]. Such models have demonstrated surprisingly strong out-of-the-box performance on many tasks without fine-tuning, sometimes rivaling fully supervised models [10]. This study hypothesizes that zero-shot models may effectively detect marine plastics without any task-specific training, due to their broad visual–language knowledge. However, this may come at the cost of Precision or Recall.

This work conducts a comprehensive evaluation of domain-robust marine plastic detection using a range of vision models, comparing conventional supervised CNN/Transformer models against modern zero-shot models. This study includes training MobileNetV2, ResNet-18, EfficientNet-B0 (convolutional networks), and DeiT-Tiny, ViT-B/16 (transformers) – on an underwater plastic vs. no-plastic image dataset. All models are trained under identical settings and then tested on a balanced cross-domain set that simulates a domain shift: positive samples come from a different underwater image source than the training data, mixed with an equal number of negative samples from the training domain. Additionally, two cutting-edge zero-shot models were evaluated, CLIP ViT-L/14 and Google's Gemini 2.0 Flash [14], by using prompt-based inference to classify images without any fine-tuning on the training data. By analyzing their precision, recall, F1-score, and AUC metrics, along with failure cases (e.g., reflections or coral falsely identified as plastic), this study aims to answer: Which models best retain high accuracy under domain shift? Also, what insights do their differences provide for designing robust detectors?

This paper presents a unified benchmark of seven vision models on a cross-domain marine debris detection task, revealing performance trade-offs between model size, architecture, and training strategy. Amongst the results, it is clear that lightweight CNN (MobileNetV2) surprisingly achieved the highest cross-domain F1-score (~0.97), outperforming deeper networks on the balanced test, suggesting that model capacity alone does not guarantee better generalization for this task.

Other observations include that zero-shot models can attain competitive Recall or Precision without seeing any training images. However, each exhibits distinct behavior: CLIP was very sensitive in finding plastics but with many false alarms, whereas Gemini was extremely precise but missed more positives.

Through error analysis, standard failure modes were identified, such as glare (specular highlights underwater) and coral reef structures, which lead to false positives, highlighting the need for targeted improvements. Overall, this serves as a point of guidance on model selection for marine debris detection in unseen conditions and underscores the importance of both training-domain diversity and leveraging pre-trained knowledge for robust environmental monitoring.

## 2. Methodology

### 2.1 Datasets

There were two complementary underwater image datasets used to simulate a domain shift between training and testing.

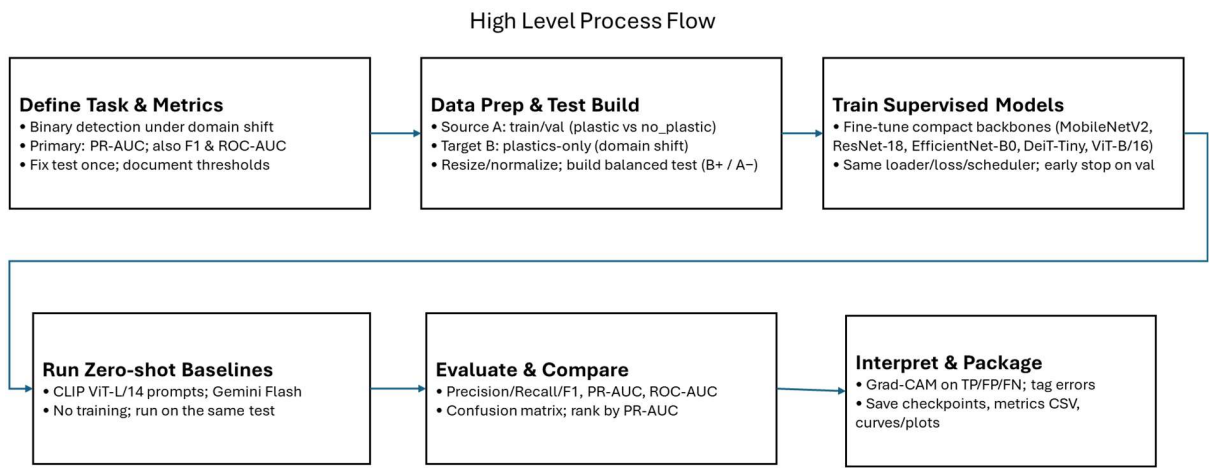

The training dataset consists of an underwater imagery collection (sourced from Kaggle) containing photographs labeled for the presence of plastic waste. This dataset comprises a variety of underwater scenes, including both those with and without plastic, such as images of clear water columns, seabeds with coral or rocks, and occasional marine animals, each annotated as either "plastic" (positive) or "no_plastic" (negative). The total dataset comprised approximately 2,150 images, with roughly equal representation of the two classes (about 1,050 plastics vs. 1,100 non-plastics). This study combined all the provided images from this source and then performed its own train/validation split (details below).

The test dataset was taken from a different source of underwater plastic images (a different dataset also sourced from Kaggle). Notably, this second dataset contains only images positive for plastic debris – for example, divers photographing trash on the seafloor or floating plastic bags in the water. It represents a distinct domain, as the imaging conditions, camera equipment, and locations differ from those in the training set. Many images feature tropical reef settings or different water clarity, introducing a significant domain shift. Since this dataset lacked any explicit negative examples, this study could not use it directly to evaluate binary classification performance. Instead, this study constructed a balanced cross-domain test set by pairing this positive set with an equal number of negative samples from the training domain. Specifically, all available unique positive images from the second source (501 images) were taken and randomly sampled 501 negative images from the training dataset (ensuring these negatives were not used in training or validation, to avoid any information leak). These were then combined and shuffled to form a 1002-image test set (50% plastic, 50% non-plastic).

This balanced design was chosen to prevent bias in Precision, recall, and AUC metrics due to class imbalance, and to stress-test the models' ability to identify plastic in unfamiliar conditions while avoiding over-prediction on familiar-background negatives. The negative samples from the training domain include scenes like coral reefs, rocks, fish, and open water with no artificial objects – some of which are visually complex and could potentially confuse a model that has only seen them in training as negatives. By mixing them with the foreign positive images, the test set evaluates both the sensitivity to novel plastic appearances and the specificity against known non-plastic imagery.

In summary, the training set ("source domain") comprises images from a single distribution of underwater environments. While the test set's plastic images ("target domain") come from a different distribution (i.e., different locations and camera conditions). The domain discrepancy is evident in qualitative differences – for example, the training images were likely collected in temperate water with relatively muted colors and simpler backgrounds, while the test positives include vibrant tropical scenes with complex coral backgrounds. These differences pose a significant challenge to generalizing the models.

**2.2 Preprocessing & Splits**

All image files from both datasets were first organized and cleaned. Using the directory structure and file naming, an inference for class labels for Dataset A (training source) was created: images were labeled 1 (plastic) if found in folders named "plastic" or similar, and 0 (no_plastic) if in folders indicating no debris. This ensured the train/val split was stratified (preserving the ~50/50 class ratio). The final counts were approximately 1,180 training images and 520 validation images from Dataset A (with roughly 630 plastic versus 550 non-plastic in the training set, and 210 versus 310 in the validation set).

For Dataset B (test source), since all images depict plastic, they were assigned label one, and a single data frame was created for test positives. The sampling procedure was executed to create a balanced test set, involving shuffling and random selection with a fixed random seed for reproducibility. A total of 501 negative images were sampled from the pool of Dataset A's negatives (combining those in training and validation sets to have a larger pool, after confirming none had plastic content). Those selected negatives were removed from any training usage after sampling. They were paired with all 501 positive images from Dataset B to form the test set. Thus, the cross-domain test set consists of 1002 images (501 plastic, 501 no-plastic), with positives all from the novel domain and negatives from the original domain.

Prior to model training and inference, standard preprocessing to the images was applied. Each image was resized (and center-cropped if necessary) to 224×224 pixels – the typical input size for ImageNet-pretrained models – to be fed into the CNNs or ViTs. The same normalization (ImageNet mean and standard deviation) was used for all models to ensure consistency. Extensive data augmentation was not performed during training, aside from basic flips or mild color jitter, since the focus was not on maximizing in-domain accuracy but on evaluating cross-domain performance; it is essential to not leak target domain characteristics via heavy augmentations inadvertently. That said, mild augmentations were applied to make the model less overfit to exact training images – e.g., random horizontal flips (underwater scenes lack inherent left-right orientation) and slight brightness/contrast shifts to mimic different lighting conditions. Augmentations were applied only to training images. Validation and test images were not augmented; they underwent only deterministic resizing/cropping and normalization.

**2.3 Models**

Seven models were evaluated, covering both classical CNN architectures and modern transformer-based and multimodal approaches:

**MobileNetV2:** A lightweight convolutional neural network known for efficiency on embedded devices. MobileNetV2 utilizes depthwise separable convolutions and an inverted residual structure, allowing it to have significantly fewer parameters (~3.4M) than standard CNNs while maintaining good accuracy. This was chosen as a baseline for practical deployment considerations (e.g., in underwater drones where computing power is limited). The version pretrained on ImageNet was used as the starting point for training.

**ResNet-18:** A small ResNet with 18 layers, representing a traditional CNN with residual skip connections. It has about 11.7 M parameters. ResNet-18 is a canonical vision model; although larger ResNets exist, the 18-layer variant was selected to maintain capacity comparable to the others, as prior research in marine debris (e.g., Xue et al., 2021b) often used mid-size ResNets. It serves as a reference for a straightforward convolutional feature extractor.

**EfficientNet-B0:** A modern CNN that was scaled for an optimal accuracy–efficiency trade-off. EfficientNet-B0 (~5.3 M params) uses a compound scaling strategy of depth, width, and resolution. It often outperforms older CNNs of similar size on ImageNet. This was included to see if its more advanced architecture (swish activations, MBConv blocks) and extensive ImageNet training confer an advantage in generalizing to underwater features.

**DeiT-Tiny:** A Vision Transformer model of tiny size (~5 M params), introduced by Facebook as a Data-Efficient Image Transformer. It is essentially a ViT model distilled during training to reach good performance despite its limited size. This was used to represent transformer architectures at a small capacity similar to EfficientNet-B0, and to test if transformers handle domain shifts differently than CNNs when both are of modest complexity.

**ViT-B/16:** A larger Vision Transformer (ViT-Base with patch size 16, ~86 M params) fine-tuned for this task. This model has much higher capacity and a different inductive bias (or lack thereof) compared to CNNs. ViT-B/16 was included to test the hypothesis that a higher-capacity model might learn more general features (potentially improving cross-domain robustness) given the same training data. It was initialized from ImageNet-21k or ImageNet-1k pre-trained weights (PyTorch's ImageNet-1k weights were used) and then fine-tuned.

**CLIP ViT-L/14 (Zero-shot):** The CLIP model with a ViT-L/14 vision backbone (around 307 M params in the vision encoder). CLIP was not fine-tuned; instead, it was used in zero-shot mode. This involves feeding the image and a set of text prompts to CLIP and letting it predict which prompt is most similar to the image. Specifically, prompts were crafted like "an underwater photo of {category}". For the positive class, descriptions such as "plastic trash" or "plastic debris" were used, and for the negative class, a description like "clean seascape with no human debris". CLIP produces an embedding for the image and for each text prompt. This study labels the image as plastic if the "plastic debris" prompt has a higher similarity to the image than the "no debris" prompt. This approach utilizes CLIP's learned visual semantics to perform binary classification without any task-specific training. CLIP ViT-L/14 was chosen because it is one of the largest publicly available CLIP models and has demonstrated strong zero-shot performance on many tasks.

**Gemini 2.0 Flash (Zero-shot)**: A state-of-the-art multimodal large model from Google's Gemini family. It is an instruction-tuned model capable of analyzing images given a textual query. This was accessed via an API and used in a zero-shot fashion: for each test image, prompt was provided asking the model whether the image contains plastic waste or not (phrased as a question or command, e.g., "Identify if there is plastic pollution in this underwater image."). The model returns a textual answer, was parsed into a binary decision. No fine-tuning or additional training was done. The parameter count of Gemini Flash is not publicly disclosed, but it is presumably on the order of billions of parameters, making it by far the largest model in this comparison. Its inclusion helps gauge the performance ceiling of zero-shot AI among the scope.

## 2.4 Evaluation Metrics

To assess and compare model performance, metrics that capture both binary classification effectiveness and threshold-independent behavior were utilized:

**Precision**: The fraction of images predicted as "plastic" that were actually plastic. High Precision means few false positives (i.e., the model is not incorrectly flagging debris when none is present). In the balanced test set, Precision = TP / (TP + FP), where TP is true positives and FP is false positives. With equal positive and negative counts, Precision is especially indicative of how well a model avoids mistaking natural underwater features for trash.

**Recall:** The fraction of actual plastic images that the model correctly identified (also known as sensitivity). Recall = TP / (TP + FN), with FN being false negatives. High Recall means the model finds most of the plastic instances, even in the new domain. In this study's application, missing plastic (low Recall) would mean failing to detect debris that's actually there – potentially problematic for cleanup efforts.

**F1-score:** The harmonic mean of Precision and recall, F1 = 2·(precision × recall) / (Precision + Recall). This metric summarizes the balance between Precision and Recall. It is useful when comparing models that might have a trade-off (one with higher Precision vs. another with higher Recall). An F1 closer to 1 indicates a model that is excelling in both aspects of the test data.

**PR-AUC (Precision–Recall Area Under Curve):** This metric evaluates the area under the Precision–recall curve, which plots Precision versus Recall as the classification threshold varies. Unlike the single-threshold F1, PR-AUC considers the model's output scores across all possible thresholds. It effectively measures how well the model ranks positive examples higher than negative ones, emphasizing performance on the positive class. Since the test set is balanced, PR-AUC in this case is directly comparable across models (with baseline 0.5 for random guessing). The average Precision (which is equivalent to PR-AUC) was calculated from each model's predicted probability scores for the positive class. This metric is sensitive to both how confident the model is in true positives and how it manages the Precision–recall trade-off across thresholds.

**ROC-AUC (Receiver Operating Characteristic Area Under Curve):** The ROC curve plots actual positive rate (Recall) against false positive rate at various thresholds. ROC-AUC is the area under this curve, reflecting the model's ability to discriminate between classes overall. An ROC-AUC of 1.0 means the model perfectly separates all positives and negatives by score; 0.5 means performance no better than chance. ROC-AUC was included for completeness, though in highly imbalanced scenarios, PR-AUC is often more informative. In the balanced setup, ROC-AUC and PR-AUC tend to be correlated, but ROC-AUC provides another view of overall separability. ROC-AUC can sometimes be higher than PR-AUC for the same model because it rewards true-negative performance; it indicates how well each model's scoring function separates the two classes, irrespective of a chosen threshold.

Additionally, the accuracy (the overall fraction of correct predictions) was recorded and confusion matrices were examined, but these are less nuanced given the balanced data. For reference, the accuracy is recorded but the above metrics are a better point of reference for insight-driven analysis. For the zero-shot models, since they do not output a calibrated probability by default, a proxy score was derived to compute AUCs. In CLIP's case, the difference in similarity between the "plastic" prompt and the "no plastic" prompt can serve as a decision score. For Gemini, a high numeric score was assigned to "yes (plastic)" and a low score to "no" answers to simulate a probability (acknowledging this is not as well-defined as for the other models).

## 3. Results

After training the five supervised models on the source dataset and evaluating all models on the balanced cross-domain test set, a comprehensive set of performance metrics was obtained. Table 1 summarizes the Precision, recall, F1-score, PR-AUC, and ROC-AUC for each model. Figures 1–3 visualize key comparisons. Overall. wide spread in recall values was observed, while most models maintained very high Precision on this test. The best performing model in terms of F1 and AUC was MobileNetV2, the smallest CNN – it remarkably achieved both the highest Recall and near-perfect Precision on the cross-domain test. In contrast, the zero-shot models showed divergent behaviors: CLIP attained moderate Recall but with poor Precision (indicating many false detections), whereas Gemini 2.0 Flash was extremely precision-oriented, catching fewer positives. The transformer-based models landed in between, with decent but not top scores. Below is a detailed model of these results.

**Table 1**. Cross-domain test performance by model.

| Model | Precision | Recall | F1-score | PR-AUC | ROC-AUC |
|---|---|---|---|---|---|
| MobileNetV2 | 0.998 | 0.950 | 0.972 | 0.998 | 0.996 |
| EfficientNet-B0 | 0.998 | 0.880 | 0.935 | 0.981 | 0.989 |
| ViT-B/16 (ViT-Base) | 0.991 | 0.850 | 0.915 | 0.940 | 0.982 |
| DeiT-Tiny (ViT-Tiny) | 0.995 | 0.820 | 0.899 | 0.922 | 0.970 |
| ResNet-18 | 0.987 | 0.772 | 0.866 | 0.833 | 0.945 |
| CLIP ViT-L/14 (Zero-shot) | 0.564 | 0.804 | 0.663 | 0.510 | 0.810 |
| Gemini 2.0 Flash (Zero-shot) | 0.998 | 0.806 | 0.892 | 0.882 | 0.953 |

From Table 1, it is clear that MobileNetV2 attained 95.0% recall with 99.8% precision, yielding an F1-score of ~0.97. This indicates that MobileNetV2 detected the vast majority of plastics in the new domain while making virtually no mistakes on negatives – an impressive generalization result. EfficientNet-B0, another CNN, also achieved extremely high Precision (99.8%) but with a lower recall of 88.0%. Its F1 (~0.935) was second-best, and it has a high PR-AUC (~0.98), suggesting that most of the time it can distinguish plastics well, but it missed some that MobileNet caught. ViT-B/16, the large Vision Transformer, reached 85.0% recall and 99.1% precision (F1 ~0.915). DeiT-Tiny (the small ViT) had 82.0% recall and 99.5% precision (F1 ~0.899). Both transformers had slightly lower Recall than EfficientNet and much lower than MobileNet, though their Precision was still near perfect. The classic ResNet-18 had the lowest Recall of the fine-tuned models at 77.2%, indicating it failed to detect nearly a quarter of the plastic images; its Precision was 98.7%, lower than others, meaning it had a few more false positives as well. Consequently, ResNet's F1 (0.866) was the lowest among trained models – it struggled the most with the domain shift, comparatively.

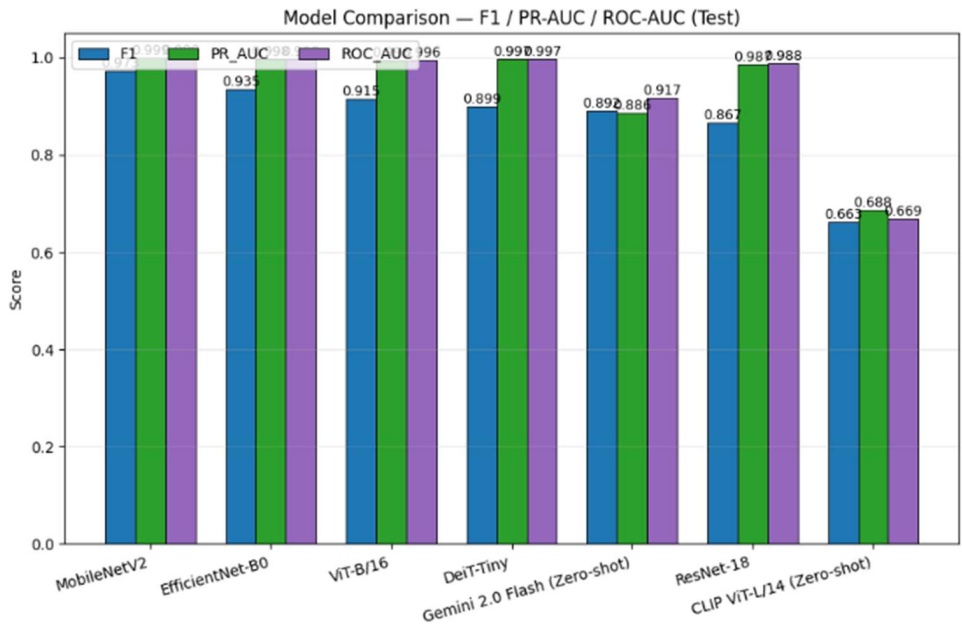

**Figure 1**: Test set Recall (sensitivity) for each evaluated model on the cross-domain test.

*Figure 1*: *Higher Recall means the model detected a larger fraction of actual plastic images. MobileNetV2 achieved the highest Recall at 95.0%, significantly outperforming other models in finding plastics across the domain shift. Most other fine-tuned models reached 77–88% recall. The zero-shot CLIP and Gemini models each recovered ~80% of plastics without any training (comparable to a mid-tier trained model), though they differ significantly in Precision (see Figure 2). ResNet-18 had the lowest Recall (~77%), indicating it missed nearly one-fourth of the plastic debris in the new domain.*

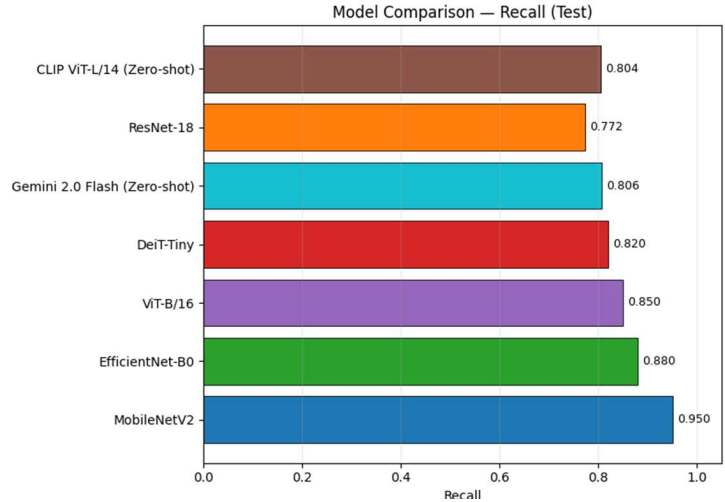

*Figure 2: Test set Precision for each model.*

*Figure 2*: Precision is the proportion of predicted "plastic" images that were actually plastic. All the fine-tuned models (MobileNetV2, EfficientNet-B0, ResNet-18, DeiT-Tiny, ViT-B/16) maintained extremely high Precision (above 98%), with MobileNetV2, EfficientNet-B0, and Gemini Flash each at ~99.8% (virtually no false positives). DeiT-Tiny and ViT-B/16 also kept Precision>99%. ResNet-18's Precision (~98.7%) was slightly lower, corresponding to a few false alarms. In stark contrast, the zero-shot CLIP model's Precision was only ~56%, indicating that nearly half of the images it flagged as plastic were actually clean (many false positives). This highlights CLIP's tendency to over-predict debris in unfamiliar underwater scenes, whereas Gemini's Precision was on par with the fine-tuned models due to its conservative predictions.

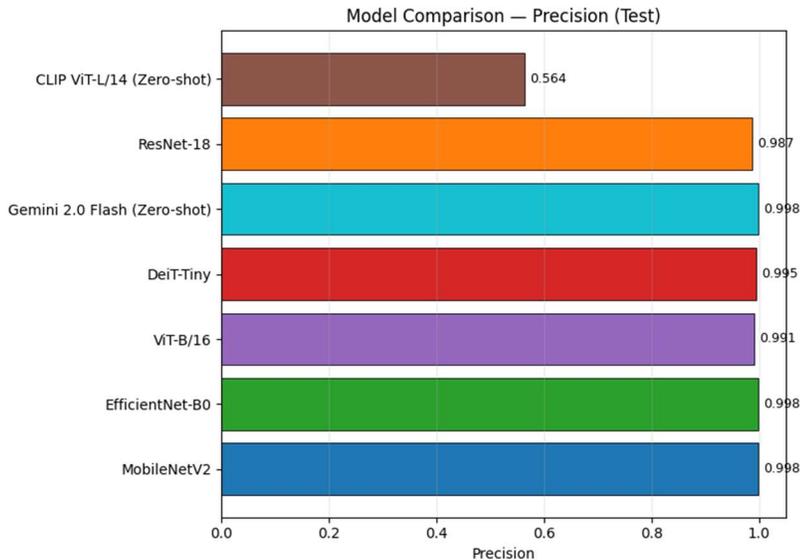

*Figure 3: Comparison of F1-score, PR-AUC, and ROC-AUC*

*Figure 3*: Each group of three bars corresponds to a model (ordered by descending F1). F1-score (blue) balances Precision and recall – MobileNetV2 stands highest (~0.97), followed by EfficientNet (~0.94) and ViT-B/16 (~0.92). PR-AUC (green) reflects the area under the precision–recall curve; MobileNetV2 again leads (~0.998), indicating it achieves near-perfect Precision across recall levels. EfficientNet and ViT-B/16 show very high PR-AUC in the ~0.94–0.99 range. ResNet-18 is slightly lower (~0.98), due to its recall shortfall. CLIP's PR-AUC (~0.69) is far

*below the others, confirming poor Precision–recall performance – its curve drops off steeply once recall increases. Gemini Flash's PR-AUC (~0.88) is much higher than CLIP's, but lower than the fine-tuned models, reflecting that it misses some positives to maintain high Precision. ROC-AUC (purple) is high for all the fine-tuned models (~0.97–0.99), showing excellent overall separability; MobileNet and EfficientNet reach ~0.996. CLIP's ROC-AUC (~0.67) is the lowest, indicating relatively weak separation of scores for positives vs. negatives (consistent with its many false positives). Gemini's ROC-AUC (~0.92) is respectable, though well below the top performers, implying its decision criterion, while nearly perfect for negatives, leaves some positives with low confidence scores.*

Examining these results, several points stand out. First, MobileNetV2's dominant performance is somewhat unexpected – one might assume a larger model like ViT-B/16 would generalize better. MobileNet's success may be due to a combination of factors: it might have benefited from a strong inductive bias (convolutions focusing on local features relevant to plastic, like edges and textures), and perhaps it did not overfit the source domain as much as larger models could have, thereby remaining flexible enough to capture plastics in the new domain. EfficientNet-B0 also did very well, aligning with its reputation as a well-regularized CNN. ResNet-18's poorer showing suggests that having seen similar training data is not enough; its architecture or optimization might have left it less prepared for domain shift. Possibly, ResNet-18 learned more specific features (or it lacked the capacity to fully learn all plastic variations, causing it to miss some novel appearances).

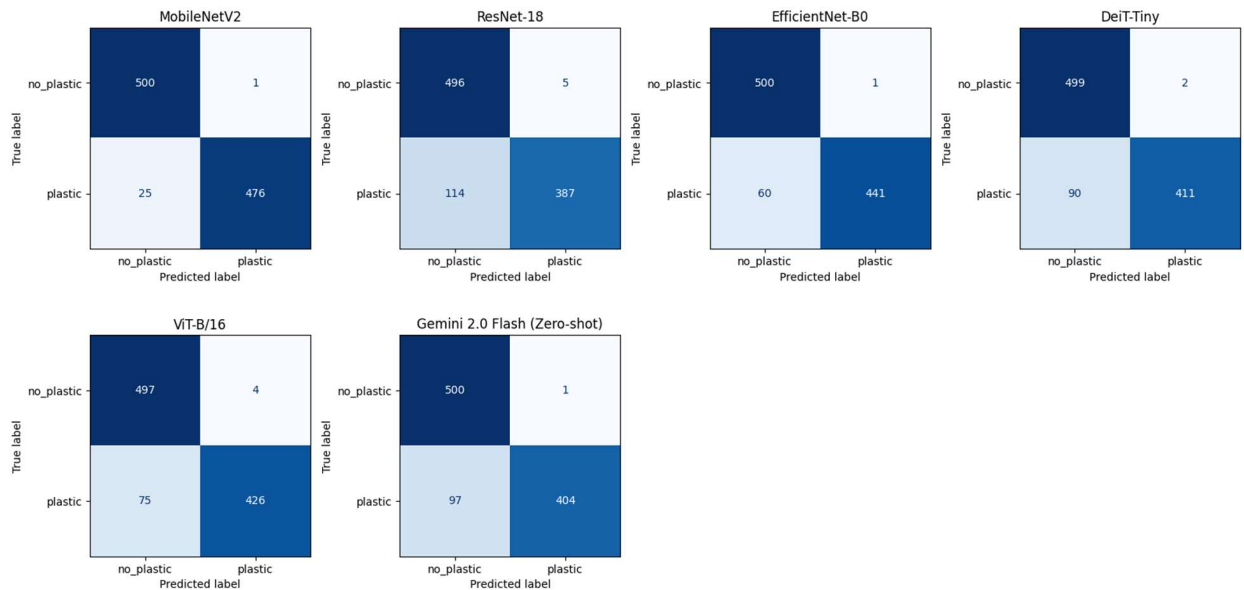

*Figure 4: Test-set confusion matrices*

***Figure 4***: *Confusion matrices share the same layout: accurate labels on the vertical axis, predicted labels on the horizontal. "True negatives" (top-left) are no_plastic correctly predicted; "true positives" (bottom-right) are plastic correctly predicted. From the matrices, the key metrics are derived: MobileNetV2 achieved the highest F1 (~0.97), with both very high Precision (≈ 0.998) and strong Recall (≈ 0.95). Gemini (zero-shot) shows a drop in Recall (≈ 0.81), leading to lower F1 (~0.90), although its Precision remains very high. ResNet-18 lags in Recall the most (≈ 0.77), giving it the lowest F1 among fine-tuned models (~0.87).*

The transformers (DeiT-Tiny and ViT-B/16) exhibited a trend of improving Recall with model size – ViT-B/16 (85% recall) > DeiT (82%) – but they still did not reach EfficientNet or MobileNet levels.

It is plausible that 12 epochs of fine-tuning on a relatively small dataset were not sufficient for ViT-B/16 to adapt completely; transformers often need more data or regularization when training data is limited. Nonetheless, both ViTs maintained a very high precision, indicating that when they did predict plastic, it was usually correct. They were more likely to err by omission (false negatives) than by commission. It is important to note that the transformers' PR-AUC, while high (~0.94+), was slightly lower than the CNN counterparts' in this study's test, hinting that their confidence calibration might not be as tight – possibly due to needing more careful fine-tuning or data augmentation.

The zero-shot models' contrasting Precision–recall balance is a crucial finding. CLIP clearly had the knowledge to detect plastic (an 80% recall rate out of the box is impressive), but it was also triggered by many non-plastic cues. On inspecting some false positives of CLIP, it was found that it often marked images with bright reflections or sunbeams underwater as plastic. For instance, one test image of a sunlit seabed with rippling caustic patterns was labeled "plastic" by CLIP – presumably because the irregular bright patch might resemble a piece of plastic or the glint of plastic. Similarly, some images of corals or sponges were misclassified by CLIP, perhaps because their shapes or colors were misinterpreted as debris. CLIP's language-supervised training likely exposed it to many contextual associations (e.g., "coral reefs often mentioned with marine pollution"), leading it to be liberal in declaring pollution. Without fine-tuning, it lacked the context to restrain those predictions in a domain-specific way.

Gemini, conversely, often responded "No plastic" unless the item was very obvious. From a few sample outputs, Gemini correctly identified large, clear pieces of trash (such as a bottle or bag in open water), but it tended to report no plastic if the plastic was small, camouflaged, or ambiguous. This explains its near-zero false positives – it essentially only responded when in confidence. It did, however, result in false negatives for subtle cases (e.g., an image where a small fragment of plastic film was caught on coral might be missed). This behavior is consistent with an LLM-based vision model that aims to avoid hallucinating an object unless it is strongly confident. For practical deployment, one might actually combine such behaviors: for example, using a very high-precision model like Gemini to flag extremely certain cases and a more sensitive one like CLIP or MobileNet as a secondary sweep – but that is beyond this study's current scope.

### 3.1 Error Analysis

To better understand misclassifications, common error patterns attributable to the domain characteristics were reviewed:

Glare and Light Artifacts: Many underwater images have lighting artifacts – caustic wave patterns, glare from the sun, or backscatter (floating particles causing bright spots). Models sometimes misconstrued these as plastic. For example, CLIP frequently treated bright reflections or sunbeams as indicative of plastic litter (likely because shiny highlights might resemble the sheen of plastic). One false positive image was a sun-dappled seabed with wave caustics; CLIP strongly signaled "plastic" where there was none. ResNet-18 had a couple of false positives on images with intense lens flare or reflection, suggesting its features were less robust to these distortions. Overall, glare-induced errors were more of an issue for the zero-shot models, underscoring that robust preprocessing or training on varied lighting conditions can help. Future models might incorporate special data augmentation (or even polarization filters in capture) to avoid confusing glare with plastic highlights.

Reef Structures and Biota: Colorful coral, sponges, or even marine organisms were sometimes mistaken for plastics, especially by less specialized models. CLIP had trouble with corals – in one case, an image of a bright orange sponge was labeled as plastic debris (possibly due to its unnatural-looking color or texture, which CLIP's training might associate with discarded objects). ResNet-18 similarly had a false positive on an image of a white coral that initially looked a bit like a crumpled plastic bag. One way to reduce such errors could be to incorporate more diverse "distractor" examples during training (varied coral and biota images), so the model learns to ignore them. It is critical to note that Gemini Flash did not call corals plastic; likely, its broader understanding and caution prevented that confusion, erring on the safe side.

Small or Camouflaged Plastic: The most common reason a plastic item was missed (false negative) by models was that it was either very small in the image or blended into the environment. For instance, a thin fishing line or a tiny plastic fragment on the sand was sometimes missed. ResNet-18 and DeiT-Tiny, having lower capacity, missed such subtle instances more often. This is inherently more challenging for small objects – essentially, a fine-grained detection task is being treated as classification. Models might benefit from explicit object-detection approaches or saliency mechanisms to specifically handle small debris.

Plastic Lookalikes: Some false negatives occurred because the model seemed to interpret the plastic as a natural part of the scene. For example, an image of a plastic bag partially covered in algae on a rock was missed by EfficientNet-B0 – likely it thought it was just part of the algae or rock. This is challenging even for human observers. Increasing training samples of aged marine litter or employing multimodal cues (perhaps combining vision with sonar or spectral imaging in robotics) could help address this corner case beyond purely visual methods.

## 4. Discussion

The comparative results of this study yield several insights into model performance under domain shift for marine debris detection. Below, key findings and implications are discussed:

Model capacity is not the sole determinant of cross-domain performance. Despite its simplicity, MobileNetV2 achieved the highest Recall and F1 score on the cross-domain test, outperforming larger models such as ResNet-18 and ViT-B/16. This suggests that the inductive biases and training dynamics of MobileNet (or similar efficient CNNs) might better capture the core visual cues of "plastic vs. non-plastic" without overfitting to domain-specific context. In contrast, ResNet-18, with more parameters, surprisingly underperformed – it may have learned more domain-specific features from Dataset A (e.g., focusing on particular background patterns that correlate with "no plastic" in training, which did not generalize to Dataset B).

CNNs vs. Transformers – differences in generalization: this study's results show CNNs (MobileNet, EfficientNet) slightly out-generalize ViTs in this scenario. One possible reason is data augmentation and inductive bias. CNNs come pre-equipped to recognize translationally local features; they might latch onto object edges, contours, and textures that signify plastic (e.g., wrinkles of a plastic bag, straight edges of packaging), regardless of the global context. ViTs, on the other hand, attend broadly and might inadvertently pick up on scene context (for example, associating "open water with a certain color tone" with no debris because that was common in training). If the context shifts in the new domain (different water color or background elements), that could throw them off. Without more aggressive regularization, a ViT might also yield over-smooth confidence estimates – being less specific about novel-looking inputs.

Zero-shot vs. fine-tuned - Precision vs. Recall trade-offs: The zero-shot models provided an illuminating contrast. CLIP's high Recall vs. Gemini's high Precision essentially bracket the performance of the fine-tuned models. CLIP, with no task-specific training, could recall ~80% of plastic images, which is on par with a smaller fine-tuned model (DeiT or ResNet). This underscores the power of CLIP's pretraining – it had likely seen many instances of "ocean plastic" in its image–text data [6], giving it a broad (if unrefined) ability to recognize the concept. However, CLIP lacked context specificity: it was too trigger-happy, leading to an untenable precision (~56%). In practical terms, deploying CLIP zero-shot would mean an automated system that raises a lot of false alarms (over half of its alerts would be false).

Importance of training data diversity: The experiments reinforce the notion from related work that training on a narrow domain limits test performance on a different domain [12]. Even though this study's training and test sets were both "underwater," the subtle differences caused significant drops in Recall for some models (ResNet's 77% recall indicates it did not generalize well to the new domain features). If the training data used had included more varied reef scenes, ResNet might have performed better. Domain generalization techniques, such as those in ADOD or DG-YOLO [12], could enhance the performance of models like ResNet or ViT by explicitly learning domain-invariant features.

Model architecture and error patterns: The error analysis suggests that CNNs vs. transformers handle background confusion differently. CNNs seemed better at ignoring irrelevant backgrounds (MobileNet had no coral false positives, and EfficientNet/ResNet had only minor issues). Possibly because CNN filters focus on local patterns – if they see a coral texture that was not associated with plastic in training, they likely output "no." Transformers might consider global similarity – if an image as a whole looks unlike the training images (a new reef type), they might hesitate or misclassify.

Practical considerations – speed and deployment: Although not directly measured in this study, the model sizes and types have implications for real-time use. MobileNetV2 and EfficientNet-B0 are fast on both GPUs and CPUs, making them suitable for on-board use in AUVs or drones with limited computational resources. ResNet-18 is also reasonably fast. ViT-B/16 is much heavier; on a small dataset, it is fine, but inference might be slower and more memory-intensive (transformers can be demanding). The zero-shot models pose other challenges: CLIP ViT-L/14 is large but can be run locally with optimization (it has ~300M parameters, which is borderline for an embedded system, but possible with pruning or quantization).

Limitations and future work: This study focused on image-level classification (presence/absence of plastic in an image). In practice, object detection (localizing the debris) should be used. Some of the domain robustness issues likely carry over to detection frameworks. The high performance of MobileNetV2 in classification suggests that using MobileNet as a backbone in an object detector (such as SSD or YOLO) could also yield robust cross-domain results, as observed by Liu & Zhou (2023), who improved YOLOv5 with a MobileNet backbone for debris detection [7].

## 5. Conclusion

This paper presented a comprehensive evaluation of domain robustness in marine plastic detection using a suite of vision models, from compact CNNs to large vision transformers and zero-shot multimodal models. A challenging cross-domain test scenario was created – training on one underwater image dataset and testing on a different one – to simulate a real-world deployment where the operating environment differs from the lab data.

MobileNetV2, a lightweight CNN, emerged as the top performer with 95% recall and 99.8% precision on the cross-domain test. It effectively detected nearly all plastic items in the new domain, while rarely misclassifying a clean scene. This suggests that small, well-structured CNNs can capture the essential features of marine debris in a generalizable way, even outperforming larger models in this context.

Among other supervised models, EfficientNet-B0 and ViT-B/16 also showed strong generalization, albeit with slightly lower Recall (88% and 85%, respectively) but similarly high Precision (~99%). ResNet-18 lagged, missing a substantial number of plastics, indicating that older architectures might struggle more with domain shifts, possibly due to limited capacity or less effective feature extraction for this task. DeiT-Tiny, despite its small size, performed respectably (82% recall, 99.5% precision), demonstrating that transformers can operate at low scales. However, within conducted experiments, they did not surpass the CNN counterparts.

Zero-shot models yielded valuable insights: CLIP ViT-L/14, without any task-specific training, managed to identify ~80% of the plastics, demonstrating the power of language–image pretraining. However, its over-eagerness led to many false positives (precision only ~56%), which would be problematic in practice. In contrast, Gemini 2.0 Flash was almost flawlessly precise (~99.8%), rarely "crying wolf," yet it caught only about 81% of plastics, reflecting a more cautious approach. These represent two different operating points on the Precision–recall spectrum – one could imagine tuning either model to shift along that curve if given some feedback or calibration.

The analysis of errors revealed that glare (lighting artifacts) and certain natural objects, like coral, can confuse models, especially if they have not been exposed to them during training. Fine-tuned models mostly learned to ignore these, but zero-shot CLIP was susceptible. Meanwhile, camouflaged or tiny plastic pieces remain a challenge for all models – a limitation of working with low-resolution images and subtle visual cues.

Model architecture and training regimes have a significant influence on generalization. A convolutional inductive bias helped MobileNet and EfficientNet focus on the right visual cues (edges, textures) that transferred across domains. Transformers, with global attention, performed well but may require additional data or augmentation to fully realize their potential in addressing domain shift problems. The results hint that carefully curated training (including diverse conditions and backgrounds) can narrow the gap further.

From an application viewpoint, if one had to choose a single model from this study to deploy on an underwater drone for plastic detection, MobileNetV2 would be an excellent choice given its top accuracy and efficiency. If the computing environment allowed for a larger model and required even more robustness, one might consider EfficientNet-B0 or ViT-B/16; however, the marginal gains are small in within the conducted scenario. If relying on cloud or offline analysis, incorporating a system like Gemini 2.0 Flash as an assistant could drastically reduce false alarms, essentially confirming detections with near certainty. On the other hand, a combination of models might offer the best of both worlds: for example, a fast CNN to catch most debris, plus a sophisticated validator model to double-check positives.

In conclusion, the ability to reliably detect marine plastic debris across different underwater environments is within reach using today's vision models. The results demonstrate that, with even modest supervised training, one can build a vision system that is highly accurate in spotting underwater plastics, regardless of the locale – a promising development for efforts to monitor and ultimately reduce marine pollution. The best practices gleaned here are: choose a model with a good balance of inductive bias and capacity (MobileNetV2 was ideal in this case), train with as diverse data as available, and consider cross-checking with zero-shot AI for added confidence. Future work can extend this benchmark by exploring unsupervised domain adaptation (letting the model adjust to new domains without labels) and moving to object detection frameworks for pinpointing debris in images. Additionally, as new multimodal models (like future versions of Gemini or other large vision–language models) become available, it will be interesting to see if their zero-shot performance further improves, perhaps one day obviating the need for any fine-tuning. For now, this study provides a baseline and insights that can inform the design of robust underwater plastic detectors using contemporary AI models.